\input{styles/header.sty}

\author{
Hualie Jiang$^{1,2}$,
Ziyang Song$^{1,2}$,
Zhiqiang Lou$^{1,2}$,
Rui Xu$^{1,2}$\thanks{Corresponding author.},
Minglang Tan$^{2}$
}
\affiliations{
$^1$Antigravity Tech \quad
$^2$Insta360 Research\\
Project page: \url{https://antigravity-tech.github.io/DA360}
}

\title{Depth Anything in 360$^\circ$: Towards Scale Invariance in the Wild}

\makeatletter
\g@addto@macro\@maketitle{%
  \begin{center}
    \captionsetup{type=figure}
    \includegraphics[width=0.86\linewidth]{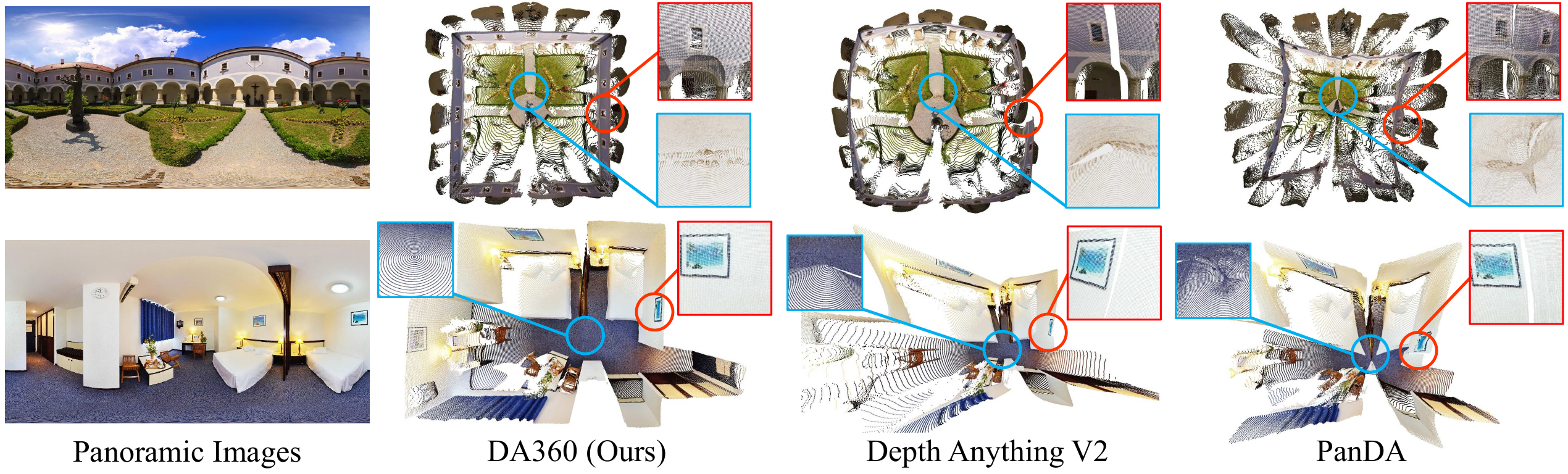}
    \captionof{figure}{Our DA360 estimates accurate scale-invariant disparity from panoramic images, which can be directly converted into well-structured 3D point clouds without artifacts at the left-right boundaries and top-down poles. In contrast, neither Depth Anything V2 (predicting affine-invariant disparity) nor PanDA (predicting affine-invariant depth) achieves this capability.}
    \label{fig:teaser}
  \end{center}%
}
\makeatother

\begin{document}

\maketitle

\begin{abstract}
Panoramic depth estimation captures the complete 360$^\circ$ scene geometry, being essential for robotics and AR/VR applications. While perspective depth models have achieved remarkable zero-shot generalization via large-scale training, panoramic methods lag behind, especially for open-world scenes, due to data scarcity. To bridge this gap, we introduce DA360, a panoramic-adapted version of Depth Anything V2. Our key insight is that the base DAV2 model, trained on perspective images to predict affine-invariant disparity, already exhibits good zero-shot performance on panoramas. Building on this, we design a lightweight adaptation framework that (i) learns a per-image shift from the ViT class token with scale-invariant supervision, transforming affine-invariant disparity into scale-invariant disparity that directly yields well-formed 3D point clouds, and (ii) integrates circular padding into the DPT decoder to eliminate seam artifacts, ensuring spatial coherence. Fine-tuned on a combination of synthetic indoor and outdoor panoramic data, DA360 is evaluated on standard real-world indoor benchmarks and our newly curated outdoor dataset, Metropolis. Results show that DA360 not only outperforms the original DAV2 by over 50\% and 12\% relative error reduction indoors and outdoors, but also surpasses prior specialized methods like PanDA by about 25--35\% across all tests, establishing state-of-the-art zero-shot panoramic depth estimation.

\end{abstract}

\section{Introduction}
\label{sec:intro}

Panoramic depth estimation, which captures 3D environmental structure from 360$^\circ$ imagery, is essential for robotics and AR/VR applications~\cite{ai2025survey}. Unlike pinhole images, panoramas provide omnidirectional perception, enabling global path planning in autonomous systems and immersive user experiences. While perspective depth estimation has achieved remarkable zero-shot generalization through large-scale training~\cite{ranftl2020towards, ranftl2021vision, depthanything, yang2024depth, ke2024repurposing, ke2025marigold}, panoramic methods remain significantly behind, primarily due to the scarcity of diverse training data. 

Existing panoramic depth estimation research has primarily focused on indoor environments, employing models trained on such constrained scenes~\cite{zioulis2018omnidepth, wang2020bifuse, jiang2021unifuse, li2022omnifusion, yun2023egformer, ai2024elite360d, lee2025hush}. However, these models struggle to generalize robustly under the domain shifts encountered in diverse, open-world settings. Meanwhile, large-scale perspective depth models such as Depth Anything~\cite{depthanything,yang2024depth} and Marigold~\cite{ke2024repurposing,ke2025marigold} exhibit strong zero-shot ability, motivating the transfer of their capabilities to the panoramic domain. Initial explorations in this direction include Depth-Anywhere~\cite{wang2024depth}, which distills pseudo-labels from Depth Anything into panoramic models~\cite{jiang2021unifuse,wang2022bifuse++}; PanDA~\cite{cao2025panda}, which directly fine-tunes the metric-depth version of Depth Anything V2 (DAV2); and DreamCube~\cite{huang2025dreamcube}, which enables seam-consistent transfer of generative models such as Marigold via cube-map synchronization.

Despite these efforts, several limitations persist. First, the choice of base model is often made without systematic investigation, limiting adaptation effectiveness. Second, these methods produce affine-invariant outputs that require post-hoc adjustment for 3D reconstruction, preventing direct conversion to well-formed point clouds. Third, their evaluations are confined to indoor benchmarks~\cite{matterport3d, armeni2017joint}, leaving outdoor generalization unverified owing to the lack of outdoor panoramic depth benchmarks.

To address these limitations, we present DA360, a framework that effectively adapts DAV2 to panoramic depth estimation. 
We begin with a comparative study of representative pinhole foundation models on panoramic data, which reveals that the base affine-invariant DAV2 exhibits the smallest domain gap. 
Building on this foundation, we introduce two lightweight adaptations: (i) a learnable shift module that, combined with a tailored scale-invariant loss, transforms affine-invariant disparity into scale-invariant disparity, removing the per-image shift ambiguity so that a single global scale factor suffices to convert the prediction into a well-formed 3D point cloud; and (ii) the complete integration into a modern DPT architecture of the circular padding that preserves ERP boundary continuity. 
To enable outdoor evaluation, we contribute Metropolis, a curated urban benchmark of 3{,}000 diverse panoramas. Experiments on indoor and outdoor benchmarks show that DA360 establishes new state-of-the-art performance, significantly outperforming prior methods.

In summary, our contributions are threefold:
\begin{itemize}
    \item 
    We identify, through a comparative study of typical depth foundation models, that the base DAV2 provides an effective starting point for panoramic adaptation.
    \item We introduce a lightweight adaptation framework centered on a learnable per-image shift with a scale-invariant loss that recovers scale-invariant disparity, enabling direct 3D point cloud generation; and the integration of circular padding into the DPT decoder to ensure spatial continuity.
    \item We contribute Metropolis, a curated set of 3{,}000 urban outdoor panoramas that fills the long-missing real-world benchmark for panoramic depth estimation, and demonstrate that DA360 achieves state-of-the-art zero-shot performance on both indoor and outdoor benchmarks.
\end{itemize}

\section{Related Work}
\label{sec:rw}

\begin{figure}[t]
\centering
\input{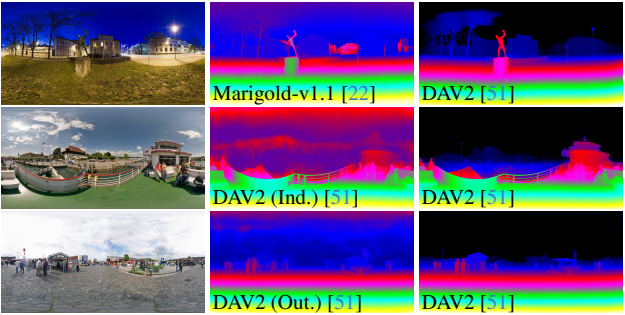}
\caption{Comparisons with SOTA zero-shot monocular depth models on outdoor panoramic images of SUN360.}
\label{fig:test_pinhole}
\end{figure}

Panoramic depth estimation extends dense geometry recovery to full 360° scenes beyond the limited FoV. The early work OmniDepth~\cite{zioulis2018omnidepth} pioneered this task, while subsequent approaches primarily address spherical distortions along two directions.
The first explores alternative projections: BiFuse~\cite{wang2020bifuse, wang2022bifuse++} fuses cubemap and ERP features through dual-branch architectures, later simplified by UniFuse~\cite{jiang2021unifuse}; OmniFusion~\cite{li2022omnifusion} processes tangent patches with geometry-aware fusion, while HRDFuse~\cite{ai2023hrdfuse} combines CNNs and Transformers for holistic-regional learning.
The second develops distortion-resistant architectures: PanoFormer~\cite{shen2022panoformer} and EGFormer~\cite{yun2023egformer} introduce distortion-aware Transformers, whereas Elite360D~\cite{ai2024elite360d} and HUSH~\cite{lee2025hush} adopt spherical representations to mitigate polar distortions.
Despite the advances, most existing methods still struggle to generalize to unseen domains.

Recent years have witnessed remarkable progress in zero-shot perspective depth estimation. MiDaS~\cite{birkl2023midas} pioneered multi-dataset training through affine-invariant prediction, and DPT~\cite{ranftl2021vision} replaced the CNN backbone with a Vision Transformer. More recent foundation models follow two paradigms: DINOv2-based ones such as Depth Anything~\cite{depthanything,yang2024depth}, and diffusion-based ones such as Marigold~\cite{ke2024repurposing, ke2025marigold}. However, these pinhole models degrade on panoramas due to spherical distortions. Sec.~\ref{sec:eval} identifies the base DAV2 as having the smallest domain gap.

Zero-shot panoramic depth estimation similarly builds on pinhole foundations and falls into two paradigms. The first estimates depth on tangent planes of the panorama and fuses the results~\cite{rey2022360monodepth, peng2023high, wang2025moge,wang2025moge2}, but suffers from slow inference and, more critically, relies on limited local context rather than the global perception that panoramas afford. The second comprises faster, end-to-end methods with global perception, including ours. Depth Anywhere~\cite{wang2024depth} distills Depth Anything predictions on unlabeled cubemap panoramas into panoramic models~\cite{jiang2021unifuse, wang2022bifuse++} to improve robustness. DreamCube~\cite{huang2025dreamcube} applies multi-plane synchronization to the neural operators of diffusion models~\cite{rombach2022high, podellsdxl, ke2024repurposing}, transferring generative capability from pinhole to panoramic imagery and thereby enabling Marigold-style panoramic depth estimation.

\begin{figure*}[t]
\centering
\includegraphics[width=0.785\linewidth]{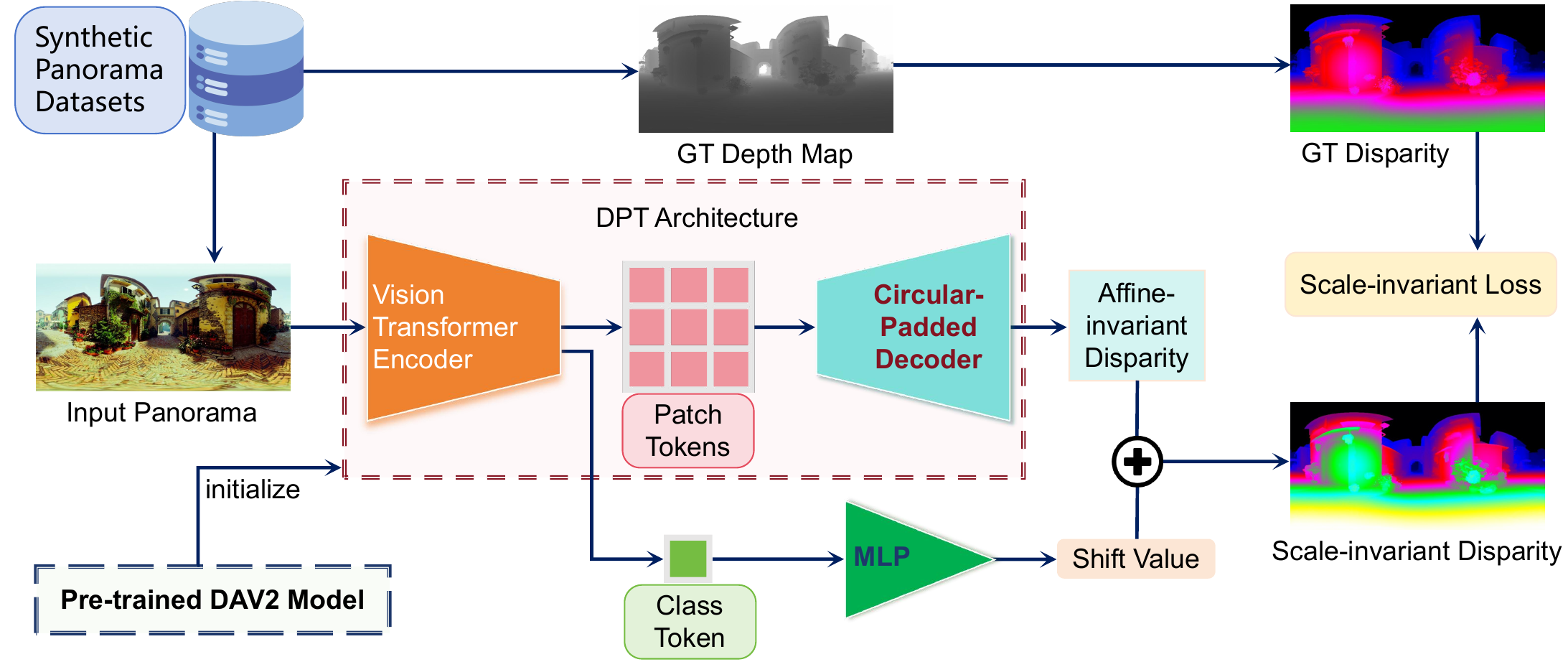}
\caption{Our framework is a generalizable panoramic depth estimation model, which fine-tunes a zero-shot pinhole depth model with synthetic panorama depth datasets to produce scale-invariant and boundary-consistent panoramic depth maps.}
\label{fig:framework}
\end{figure*}

Most closely related is PanDA~\cite{cao2025panda}, which evaluates DAV2 across five panoramic projection formats, identifies direct ERP processing as optimal, and fine-tunes DAV2 in ERP space with labeled data followed by unlabeled-data distillation. While effective, such methods typically output affine-invariant depth and have mainly been evaluated indoors. In contrast, we learn a shift for scale-invariant predictions that support direct 3D lifting, and we contribute an outdoor benchmark. We also compare with methods for arbitrary camera types: UniK3D~\cite{piccinelli2025unik3d} disentangles camera parameters from scene geometry through spherical 3D representations for metric reconstruction across camera models, while Depth Any Camera (DAC)~\cite{DepthAnyCamera} uses a pitch-aware image-to-ERP transformation trained in a unified equirectangular space. The contemporaneous DA$^{2}$~\cite{li2025depth} likewise targets scale-invariant panoramic depth, but constructs training data by outpainting six pinhole datasets and supervises only within the original FoV; we instead fine-tune a pretrained DAV2 prior more efficiently (at most 8 RTX 4090 GPUs, versus 32 H20 GPUs required by DA$^{2}$).

\section{Evaluation of Zero-shot MonoDepth Models}
\label{sec:eval}

To develop an effective zero-shot panoramic depth model, we first identify the monocular depth estimator with the smallest domain gap to the panoramic depth task. Accordingly, we evaluate two mainstream generalizable monocular depth models on the panoramic depth test sets, Matterport3D and Metropolis: Marigold, based on a diffusion model, and Depth Anything, based on the DPT architecture. For Marigold, we evaluate its updated version~\cite{ke2025marigold}. For Depth Anything, we evaluate the base DAV2~\cite{yang2024depth}, together with its fine-tuned variants for metric depth estimation in indoor and outdoor scenes, \ie, DAV2-ind and DAV2-out.

These models use distinct output types. Marigold predicts affine-invariant depth normalized to $[0,1]$. The base DAV2 follows the practice of MiDaS~\cite{ranftl2020towards} to estimate affine-invariant disparity with supervision in disparity space.
In contrast, DAV2-ind and DAV2-out target metric depth via \textit{Sigmoid} outputs scaled by 20 and 80. 
Thus, each model requires a specific alignment: affine alignment in depth space for Marigold; ground-truth inversion followed by disparity-space alignment for the base DAV2; and, similarly, affine alignment in depth space for DAV2-ind and DAV2-out, as panoramic distortion affects their depth magnitude predictions.

\begin{table}
\centering
\resizebox{0.99\columnwidth}{!}{
\begin{tabular}{l|c|cc|cc}
\toprule
\multirow{2}{*}{Models} & \multirow{2}{*}{Backbone} &\multicolumn{2}{c|}{Matterport3D} &\multicolumn{2}{c}{Metropolis}\\
& & \textit{AbsRel}$\downarrow$ & $\delta_1$$\uparrow$ & \textit{AbsRel}$\downarrow$ & $\delta_1$$\uparrow$ \\
\midrule
Marigold-v1.1~\cite{ke2025marigold}& SD 2.0 & 0.2097 & 65.11 & 0.7357 & 26.87 \\
\hline
DAV2~\cite{yang2024depth}& \multirow{3}{*}{ViT-S} & \cellcolor[gray]{.9}0.2032 &  \cellcolor[gray]{.9} 64.56  & \cellcolor[gray]{.9} 0.2169  & \cellcolor[gray]{.9} 66.34 \\
DAV2-ind~\cite{yang2024depth} & & 0.2170 &  63.98  &   0.5667 &    24.89 \\
DAV2-out~\cite{yang2024depth} & & 0.2695 &   54.18  &   0.4274 &    34.71 \\
\hline
DAV2~\cite{yang2024depth}& \multirow{3}{*}{ViT-B} & \cellcolor[gray]{.9} 0.1966&   66.03  &  \cellcolor[gray]{.9} 0.2061  & \cellcolor[gray]{.9}  67.64 \\
DAV2-ind~\cite{yang2024depth}  & & 0.2066  &  \cellcolor[gray]{.9} 66.05  &   0.6088  &   24.28 \\
DAV2-out~\cite{yang2024depth} & &  0.2382  &   59.17  &   0.4146  &   35.26 \\
\hline
DAV2~\cite{yang2024depth}& \multirow{3}{*}{ViT-L} & \cellcolor[gray]{.9}\bf{0.1926}  &   66.46  & \cellcolor[gray]{.9}\bf{0.1991}  & \cellcolor[gray]{.9}\bf{69.14} \\
DAV2-ind~\cite{yang2024depth} & &  0.2001   & \cellcolor[gray]{.9} \bf{67.47}  &   0.6193 &   24.17 \\
DAV2-out~\cite{yang2024depth} & &  0.2166   &   64.37  &   0.4193 &   34.22 \\
\bottomrule
\end{tabular}
}
\caption{Evaluating pinhole models on panorama datasets.}
\label{tab:eval_pinhole}
\end{table}

Results in Tab.~\ref{tab:eval_pinhole} show that all models perform reasonably well on the indoor Matterport3D, with DAV2-out slightly inferior. On the challenging outdoor Metropolis, performance degrades across all models, but the base DAV2 consistently achieves the lowest error and retains a smaller indoor-to-outdoor drop. DAV2-out fares better than the metric indoor variant, yet both are substantially outperformed by the base DAV2. Qualitative comparisons on SUN360 outdoor panoramas (Fig.~\ref{fig:test_pinhole}) further confirm this advantage. We attribute it to the disparity-space representation, which naturally accommodates very large distances (\eg, the sky) with compact numerical values; in contrast, depth-space models must predict extremely large magnitudes for such regions, introducing numerical instability and degrading distant-region estimates.


Based on this analysis, we select the base DAV2 as our foundation and conduct all subsequent adaptation in disparity space. This choice contrasts with prior work such as PanDA~\cite{cao2025panda}, which adopts DAV2-ind and applies affine-invariant supervision in depth space. This comparison validates disparity-space adaptation as an effective and principled starting point under a controlled, architecture-matched setting, rather than an exhaustive search for the globally optimal foundation model across all recent architectures.

\section{Adaptation Framework}
\label{sec:method}

\subsection{Overall Framework}
Fig.~\ref{fig:framework} illustrates the framework of DA360. We initialize with the base DAV2 model (based on Sec.~\ref{sec:eval}) and fine-tune it on synthetic panoramic datasets. The DPT architecture is modified with two key adaptations, each addressing a fundamental challenge identified in the introduction: (i)~achieving scale-invariant disparity, and (ii)~ensuring spatial coherence. 

\begin{figure*}[t]
\centering
\includegraphics[width=0.75\linewidth]{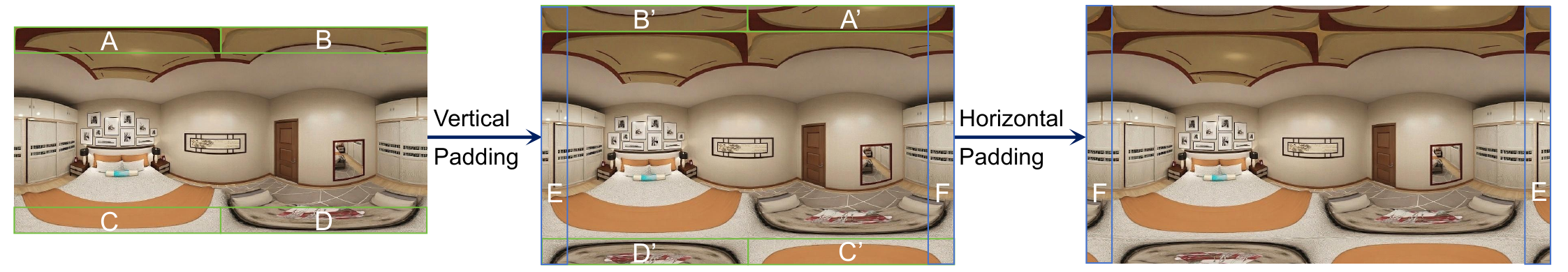}
\caption{Circular padding for equirectangular representation, divided into two phases of vertical and horizontal directions.}
\label{fig:cirp}
\end{figure*}

\subsection{Achieving Scale-Invariant Disparity}
\label{sec:si}

\textbf{Learning Shift from Class Token.}
To promote DAV2's affine-invariant disparity to a scale-invariant one, we introduce a lightweight module that predicts a global shift. The ViT backbone~\cite{dosovitskiy2020vit} learns a class token that encapsulates a holistic image representation. We leverage this token to regress the shift through a three-layer MLP, whose hidden dimensions are one-half and one-quarter of the input token dimension. The predicted shift is added element-wise to the initial disparity map produced by the DPT decoder, enabling the potential to eliminate the shift uncertainty.

\textbf{Scale-Invariant Loss.}
The shift module alone is insufficient; the network must also be supervised to predict scale-consistent disparity. We design a scale-invariant loss by modifying the affine-invariant loss from MiDaS~\cite{ranftl2020towards}. For a disparity map $\mathbf{d}$, we compute robust estimates of shift and scale over the valid pixels as:
\begin{equation}
t(\mathbf{d}) = \text{median}(\mathbf{d}), \quad s(\mathbf{d}) = \text{mean}(|\mathbf{d} - t(\mathbf{d})|).
\end{equation}
Both the predicted disparity $\mathbf{d}_{pd}$ and ground-truth disparity $\mathbf{d}_{gt}$ are normalized using \textbf{only} their respective robust scale estimates \textbf{without} the shift estimates:
\begin{equation}
\hat{\mathbf{d}}_{pd} = \frac{\mathbf{d}_{pd}}{s(\mathbf{d}_{pd})}, \quad \hat{\mathbf{d}}_{gt} = \frac{\mathbf{d}_{gt}}{s(\mathbf{d}_{gt})}.
\end{equation}
We then compute the scale-invariant loss as the L1 difference between the normalized maps:
\begin{equation}
\mathcal{L}_{si} = \frac{1}{N} \sum_{i=1}^{N} \left| \hat{\mathbf{d}}_{pd}^{(i)} - \hat{\mathbf{d}}_{gt}^{(i)} \right|.
\label{eq:si_loss}
\end{equation}
where $N$ denotes the number of valid pixels. This loss, together with the learned shift, resolves the per-image shift ambiguity, enabling the prediction to be converted into a well-formed 3D point cloud using only a global scale factor—rather than a per-image affine alignment—although metric scale remains unrecovered due to the absence of metric supervision.

\subsection{Ensuring Spatial Coherence} 
\label{sec:cpd}

While the ViT backbone captures global features via self-attention, the DPT decoder relies on convolutional layers that typically employ zero padding. This creates a critical inconsistency for ERP images: zero padding prevents features at one boundary from perceiving their physically adjacent (but ERP-separated) counterparts, leading to discontinuities along the left-right seams and artifacts in polar regions. To address this, we integrate circular padding~\cite{wang2018omnidirectional} into all convolutional layers of the DPT decoder, which is the first such integration in a modern DPT architecture for dense panoramic prediction to our knowledge.

The circular padding operation, illustrated in Fig.~\ref{fig:cirp}, proceeds in two phases. In the vertical phase, the top side of the ERP is flipped vertically, rolled horizontally by half the image width, and padded above; the bottom side is processed likewise. In the horizontal phase, the left side is padded to the right and vice versa. This lets convolutional kernels access spherical neighbors across boundaries, ensuring spatial continuity. Notably, most prior panoramic works~\cite{de2018eliminating, albanis2021pano3d, shen2022panoformer} apply circular padding only in the horizontal direction, overlooking the vertical continuity at the poles, which our two-phase procedure explicitly addresses.

\section{Datasets}
\label{sec:datasets}

\subsection{Training Sets}
\label{sec:train}

Beyond following PanDA~\cite{cao2025panda} in employing the synthetic datasets Structured3D~\cite{zheng2020structured3d} (indoor) and Deep360~\cite{li2022mode} (outdoor) as candidate training sources, we introduce TartanAir v2~\cite{patel2025tartanground}, a synthetic dataset that focuses primarily on outdoor scenes.
For Structured3D, we select 2{,}600 scenes with scene indices ${\leq}$3{,}400 for training, using both the \textit{full} and \textit{simple} style images, which totals 32{,}435 samples. For validation, we use the remaining 100 scenes and employ only the \textit{full} style images, which yields 701 samples. Deep360 is a panoramic depth dataset of driving scenarios rendered via CARLA~\cite{dosovitskiy2017carla}, containing six 500-frame episodes. We reserve the first five episodes for training and the sixth for validation. However, we observe that Deep360 erroneously renders the depth of glass surfaces (\eg, car windows) as that of the objects behind them. We thus exclude Deep360 from our final model training and use it solely in ablation studies.

TartanAir v2 provides 74 Unreal Engine scenes designed for odometry tasks. To adapt it for panoramic depth estimation, we downloaded the left-view images and depth data from the \textit{Hard} sequences and synthesized upright ERP panorama-depth pairs using the provided camera poses. We identified and removed 13 scenes that exhibit the same glass-rendering errors as Deep360, and we additionally filtered out samples with incorrect ground depth (rendered as infinite) or extreme brightness and contrast using a custom script. The final high-quality training subset comprises {432k samples from 61 scenes}. This rich and varied collection substantially improves outdoor generalization, as demonstrated by our controlled experiments in Sec.~\ref{sec:pc} and supplement (Sec.~S2.3), where PanDA retrained on the same data still lags behind DA360.

\begin{table*}
\centering
\resizebox{2.1\columnwidth}{!}{
\begin{tabular}{l|ccccc|ccccc|ccccc}
\toprule
\multirow{2}{*}{Models} &\multicolumn{5}{c|}{Matterport3D} &\multicolumn{5}{c|}{Stanford2D3D} &\multicolumn{5}{c}{Metropolis}\\
 & \textit{AbsRel}$\downarrow$ & \textit{MAE}$\downarrow$ & \textit{RMSE}$\downarrow$ & \textit{RMSE}log$\downarrow$ & $\delta_1\uparrow$ & \textit{AbsRel}$\downarrow$ & \textit{MAE}$\downarrow$ & \textit{RMSE}$\downarrow$ & \textit{RMSE}log$\downarrow$ & $\delta_1\uparrow$ & \textit{AbsRel}$\downarrow$ & \textit{MAE}$\downarrow$ & \textit{RMSE}$\downarrow$ & \textit{RMSE}log$\downarrow$ & $\delta_1\uparrow$ \\
\midrule
DreamCube~\cite{huang2025dreamcube} & 0.2736 & 0.5201 & 0.7258 & 0.1364 & 56.60 & 0.2592 & 0.4130 & 0.5773 & 0.1282 & 61.24 & 0.4928 & 15.872 & 20.682 & 0.3036 & 36.16 \\
DAC (L)~\cite{DepthAnyCamera} & 0.1442 & 0.3185 & 0.5013 & 0.0806 & 82.52 & 0.1261 & 0.2166 & 0.3495 & 0.0718 & 86.23  & 0.5524 & 17.068 & 21.748 & 0.2337 & 34.09 \\
UniK3D (L)~\cite{piccinelli2025unik3d} & 0.1046 & \cellcolor[gray]{.9}0.2133 & \cellcolor[gray]{.9}\bf{0.3906} & \cellcolor[gray]{.9}0.0645 & \cellcolor[gray]{.9}92.17 & 0.1011 & 0.1584 & {0.2706} & 0.0607 & 91.88 & 0.4842 & 15.687 & 20.847 & 0.2170 & 33.82 \\
DA$^{2}$ (L)~\cite{li2025depth} & \cellcolor[gray]{.9}0.1032 & 0.2256 & {0.4207} & 0.0658 & 89.49 & \cellcolor[gray]{.9}0.0688 & \cellcolor[gray]{.9}\bf{0.1165} & \cellcolor[gray]{.9}\bf{0.2508} & \cellcolor[gray]{.9}\bf{0.0485} & \cellcolor[gray]{.9}\bf{95.55} & \cellcolor[gray]{.9}0.3895 & \cellcolor[gray]{.9}11.658 & \cellcolor[gray]{.9}16.131 & \cellcolor[gray]{.9}0.1833 & \cellcolor[gray]{.9}45.82 \\
\midrule
DAV2 (S)~\cite{yang2024depth} & 0.2032 & 0.4553 & 0.7566 & 0.1128 & 64.56 & 0.2561 & 0.4907 & 0.7906 & 0.1416 & 55.74 & 0.2169 & 9.4807 & 16.057 & 0.1426 & 66.34 \\
PanDA (S)~\cite{cao2025panda} & 0.1293 & 0.2929 & 0.5019 & 0.0775 & 85.43 & 0.1301 & 0.2154 & 0.3530 & 0.0762 & 82.16 & 0.3045 & 11.455 & 16.773 & 0.1695 & 51.05 \\
PanDA (S)*~\cite{cao2025panda} & 0.1346 & 0.3017 & 0.5090 & 0.0784 & 84.59 & 0.0957 & 0.1694 & 0.3261 & 0.0609 & 90.80 & 0.2898 & 9.7515 & 15.037 & 0.1774 & 57.52 \\
\bf{DA360 (S)} & \cellcolor[gray]{.9}0.0963 & \cellcolor[gray]{.9}0.2364 & \cellcolor[gray]{.9}0.4971 & \cellcolor[gray]{.9}0.0682 & \cellcolor[gray]{.9}91.60 & \cellcolor[gray]{.9}0.0687 & \cellcolor[gray]{.9}0.1289 & \cellcolor[gray]{.9}0.3009 & \cellcolor[gray]{.9}{0.0532} & \cellcolor[gray]{.9}{94.84} & \cellcolor[gray]{.9}0.1922 & \cellcolor[gray]{.9}8.0552  & \cellcolor[gray]{.9}14.286 & \cellcolor[gray]{.9}0.1324 & \cellcolor[gray]{.9}74.22 \\
\hline
DAV2 (B)~\cite{yang2024depth} & 0.1966 & 0.4375 & 0.7293 & 0.1095 & 66.03 & 0.2469 & 0.4720 & 0.7595  & 0.1371 & 56.46 & 0.2061 & 9.1553 & 15.467 & 0.1371 & 67.64 \\
PanDA (B)~\cite{cao2025panda} & 0.1265 & 0.2891 & 0.4926 & 0.0754 & 86.01 & 0.1123 & 0.1907 & 0.3322 & 0.0662 & 87.49 & 0.3050 & 11.458 & 16.725 & 0.1621 & 50.40 \\
\bf{DA360 (B)} & \cellcolor[gray]{.9}0.0877 & \cellcolor[gray]{.9}0.2130 & \cellcolor[gray]{.9}0.4684 & \cellcolor[gray]{.9}0.0653 & \cellcolor[gray]{.9}92.93 & \cellcolor[gray]{.9}\bf{0.0651} & \cellcolor[gray]{.9}{0.1223} & \cellcolor[gray]{.9}0.3010 & \cellcolor[gray]{.9}0.0537 & \cellcolor[gray]{.9}94.76 & \cellcolor[gray]{.9}0.1798 & \cellcolor[gray]{.9}7.5411 & \cellcolor[gray]{.9}13.754 & \cellcolor[gray]{.9}0.1285 & \cellcolor[gray]{.9}76.68 \\
\hline
DAV2 (L)~\cite{yang2024depth} & 0.1926 & 0.4275 & 0.7066 & 0.1078 & 66.46 & 0.2595 & 0.5053 & 0.8233  & 0.1465 & 54.64 & 0.1991 & 8.9007 & 15.177  & 0.1342 & 69.14 \\
PanDA (L)~\cite{cao2025panda} & 0.1181 & 0.2707 & 0.4745 & 0.0721 & 87.61 & 0.1095 & 0.1876 & 0.3345 & 0.0667 & 87.65 & 0.2784 & 10.803 & 16.106 & 0.1537 & 54.58 \\
\bf{DA360(L)} & \cellcolor[gray]{.9}\bf{0.0793} & \cellcolor[gray]{.9}\bf{0.1931} & \cellcolor[gray]{.9}{0.4407} & \cellcolor[gray]{.9}\bf{0.0619} & \cellcolor[gray]{.9}\bf{94.07} & \cellcolor[gray]{.9}0.0710 & \cellcolor[gray]{.9}0.1327 & \cellcolor[gray]{.9}{0.2930} & \cellcolor[gray]{.9}0.0558 & \cellcolor[gray]{.9}93.53 & \cellcolor[gray]{.9}\bf{0.1630} & \cellcolor[gray]{.9}\bf{6.8220} & \cellcolor[gray]{.9}\bf{13.015} & \cellcolor[gray]{.9}\bf{0.1226} & \cellcolor[gray]{.9}\bf{80.47} \\
\bottomrule
\end{tabular}
}
\caption{Quantitative comparison. DA360 and DA$^{2}$ are evaluated with scale alignment, while all other methods use affine alignment. The postfixes (S), (B), and (L) denote the backbone scales, i.e., ViT-Small, ViT-Base, and ViT-Large, respectively.}
\label{tab:main}
\end{table*}

\subsection{Indoor Test Sets}
\label{sec:test}

Conventionally, we employ the test splits of two real-world indoor datasets: Matterport3D~\cite{matterport3d} and Stanford2D3D~\cite{armeni2017joint}, consistent with prior works~\cite{jiang2021unifuse, cao2025panda}. Both are captured using Matterport cameras but differ in scene variety. Matterport3D covers 90 diverse indoor scenes, of which 2{,}014 samples from 18 scenes are used for testing. Stanford2D3D encompasses 6 building areas from a campus, and we adopt the standard test set (Area~5), containing 373 samples.

\begin{figure}
\centering
\includegraphics[width=1.0\linewidth]{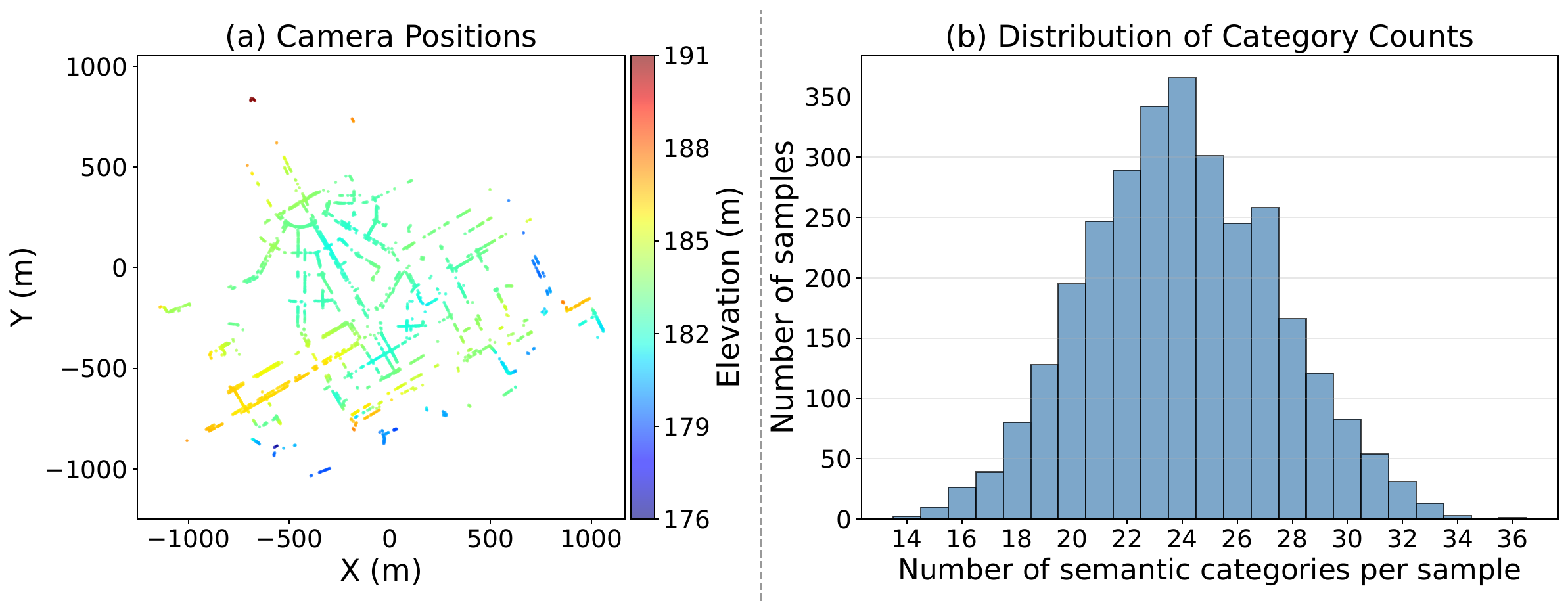}
\caption{Analysis of the curated Metropolis test set. (a) Camera positions colored by elevation, showing 2.2\,km $\times$ 1.9\,km spatial coverage in Detroit. (b) Histogram of semantic category counts per image; the average is 24 and the range is $[14, 36]$, indicating varying scene complexity.}
\label{fig:metropolis}
\end{figure}

\subsection{Metropolis: A New Outdoor Test Benchmark}
\label{sec:metro}

To fill the critical gap in open-world outdoor evaluation, we introduce \textbf{Metropolis}, a curated test set derived from the public Mapillary Metropolis dataset~\cite{mapillary_metropolis}. The original data provides ERP panoramas and cubemap depth maps generated from aerial/street-level LiDAR, SfM, and MVS. As these modalities are not perfectly aligned, we synchronized them to obtain approximately 10k candidate samples. We inpainted the bottom region of panoramas following UniFuse~\cite{jiang2021unifuse} and converted the cubemap depth to ERP depth.

\textbf{Quality Filtering.}
Observing substantial variation in depth sparsity and noise---with some samples exhibiting extremely sparse valid depths or obvious artifacts---we performed rigorous quality filtering to retain only 3{,}000 high-quality images, ensuring a more reliable test set. Full construction details are given in the supplement (Sec.~S1).

\textbf{Dataset Characteristics.} As illustrated in Fig.~\ref{fig:metropolis}(a), the curated samples span a continuous urban area in Detroit, with the positions revealing broad spatial coverage across dozens of distinct trajectories. Leveraging the dense 88-class semantic annotations, we analyzed the semantic composition of each sample. Fig.~\ref{fig:metropolis}(b) shows the distribution of category counts per image, which follows an approximately normal distribution, indicating that the dataset provides sufficient semantic richness while spanning a wide range of scene complexity. In addition, the valid depths are concentrated in content-rich mid-to-far range regions, and the images exhibit high dynamic range with natural illumination variations. These factors make Metropolis a rigorous testbed for evaluating generalization to real-world outdoor environments.

\section{Experiments}
\label{sec:exp}

\subsection{Implementation Details}
We implement our model in PyTorch and conduct all experiments on NVIDIA RTX 4090 GPUs. For training, we adopt the AdamW optimizer with a one-cycle learning rate schedule, and set the peak learning rate to $1{\times}10^{-5}$. All input images are processed at a resolution of $518{\times}1036$ pixels during both training and evaluation. For the final models, we use three ViT backbone scales (Small, Base, and Large), and train each for 300k steps with batch sizes of 32/24/16 on 2/4/8 GPUs, respectively. In ablation studies, we train the ViT-Small--based models for 100k steps with a batch size of 8 on a single GPU. Following~\cite{wang2020bifuse, jiang2021unifuse, zhuang2022acdnet}, we report \textit{AbsRel}, \textit{MAE}, \textit{RMSE}, \textit{RMSE}log, and $\delta_1$.

\begin{figure*}[t]
\centering
\input{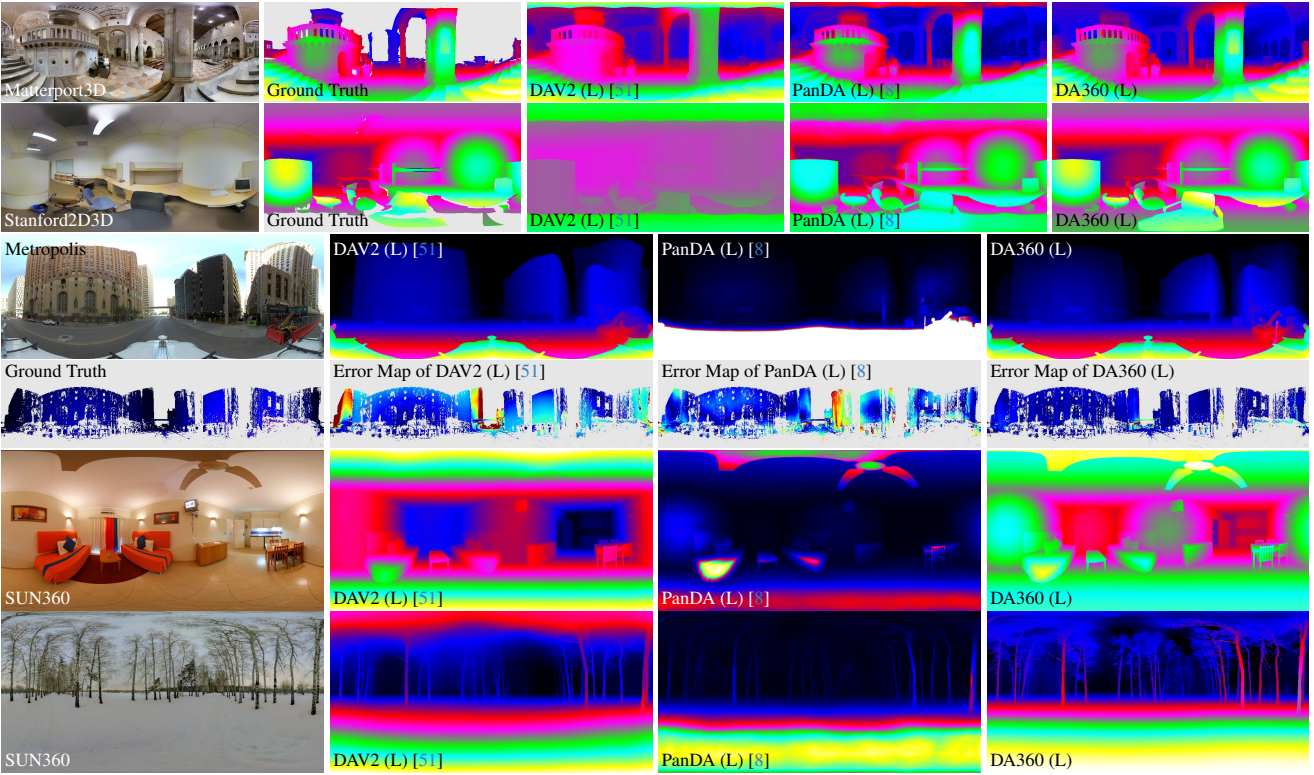}
\caption{Zero-shot qualitative comparison on Matterport3D~\cite{matterport3d}, Stanford2D3D~\cite{armeni2017joint}, Metropolis~\cite{mapillary_metropolis}, and SUN360~\cite{xiao2012recognizing}.}
\label{fig:zeroshot}
\end{figure*}

\subsection{Benchmark Comparisons}
\label{sec:pc}
This section presents qualitative and quantitative comparisons on Matterport3D~\cite{matterport3d}, Stanford2D3D~\cite{armeni2017joint}, and Metropolis~\cite{mapillary_metropolis}, with SUN360~\cite{xiao2012recognizing} included for qualitative analysis. Our comparison primarily includes the base DAV2~\cite{yang2024depth} and the closely related PanDA~\cite{cao2025panda} across all model scales; for the quantitative evaluation we further incorporate the recent end-to-end models DreamCube~\cite{huang2025dreamcube}, DAC~\cite{DepthAnyCamera}, UniK3D~\cite{piccinelli2025unik3d}, and concurrent DA$^{2}$~\cite{li2025depth} (the largest models for DAC/UniK3D; the updated Marigold~\cite{ke2025marigold} for DreamCube's depth estimation component).

Quantitative results are summarized in Tab.~\ref{tab:main}. All models perform better indoors than outdoors, which confirms that outdoor panoramic depth estimation is more challenging. DreamCube, DAC, and UniK3D exhibit poor outdoor performance (\textit{AbsRel} around 0.5), which highlights the difficulty of open-world generalization; the indoor accuracy of DreamCube is only comparable to that of the base DAV2, suggesting that diffusion models are difficult to adapt to panoramas. While DAC and UniK3D achieve competitive results indoors---with UniK3D even surpassing DA360 in \textit{RMSE}---the training data of UniK3D contains Matterport3D, and thus it cannot be regarded as zero-shot on indoor benchmarks.

Compared with the DA$^{2}$, DA360 is substantially stronger on Matterport3D and Metropolis: the \textit{RMSE} is comparable, but DA360 achieves clearly better results on the other metrics, with \textit{AbsRel} gaps of 21\% and 58\%, respectively. On Stanford2D3D, DA360 performs comparably to DA$^{2}$, although the limited scene variety of this dataset makes it less representative than Matterport3D and Metropolis. We include visual comparisons with UniK3D and DA$^{2}$ in the supplementary video, all of which demonstrate the superiority of DA360.

Compared with DAV2, DA360 shows consistent gains both indoors and outdoors: the \textit{AbsRel} decreases by over 50\% on Matterport3D and over 70\% on Stanford2D3D, while $\delta_1$ improves by over 40\% and about 70\%, respectively, and these gains hold consistently across backbone scales. On Metropolis, DA360 still achieves over 12\% gains on both \textit{AbsRel} and $\delta_1$---a smaller but substantial improvement that reflects the greater difficulty of outdoor generalization.

Compared with PanDA, DA360 reduces the \textit{AbsRel} by ${>}25\%$ on Matterport3D and ${>}28\%$ on Stanford2D3D; on Metropolis, the \textit{AbsRel} improves by ${\sim}35\%$ and $\delta_1$ by ${>}45\%$. 
PanDA remains competitive on \textit{RMSE}, likely due to its quadratic sensitivity to the slightly more high-error outliers that DA360 produces, but DA360 keeps a clear advantage in \textit{MAE} and log-\textit{RMSE}. For an even more equitable comparison, we retrain a supervised PanDA (S) on Structured3D+TartanAir v2 under identical specifications, denoted as PanDA (S)* in Tab.~\ref{tab:main}; it still trails DA360 (S) by about 28\% in \textit{AbsRel} across the three test sets, which indicates that our approach uses the training data more effectively.

Fig.~\ref{fig:zeroshot} shows qualitative results. Predictions on Matterport3D, Stanford2D3D, and Metropolis are aligned to the ground truth with the appropriate strategy for each method; for SUN360, where no ground truth exists, we use the depth of PanDA directly and invert the disparity of DA360, and we add a small bias (0.05) before inverting the disparity of DAV2 to avoid instability at zeros. DA360 recovers more accurate magnitudes and finer local details throughout. DAV2 captures the scene structure, but its global depth scale can be highly inaccurate (\eg, Stanford2D3D). On Metropolis, where valid depths concentrate in low ERP regions, the aligned predictions of PanDA appear over-compressed nearby---an issue that the quantitative metrics do not reflect. On SUN360, without ground-truth alignment, PanDA again shows unrealistic depth contrast, while DA360 remains visually consistent.

\begin{table*}[t]
\centering
\resizebox{2.08\columnwidth}{!}{
\begin{tabular}{l|c|c|c|c|c|c|cc|cc|cc|cc}
\toprule
Model & Learnable & Learnable & Circular & Training & \multirow{2}{*}{Loss} & Evaluation &\multicolumn{2}{c|}{Structured3D} &\multicolumn{2}{c|}{Deep360} &\multicolumn{2}{c|}{Matterport3D} &\multicolumn{2}{c}{Metropolis}\\
Names &  Shift & Scale & Padding & Datasets & & Alignment & \textit{AbsRel}$\downarrow$ & $\delta_1\uparrow$ & \textit{AbsRel}$\downarrow$ & $\delta_1\uparrow$ & \textit{AbsRel}$\downarrow$ & $\delta_1\uparrow$ & \textit{AbsRel}$\downarrow$ & $\delta_1\uparrow$ \\
\midrule
\textit{SI} & \checkmark & & & S+T & SI & scale & 0.0263 & 99.07 & 0.0879 & 89.34 & 0.1004 & 91.02 & 0.1966 & 72.78 \\
\hline
\textit{NoShift} & & & & S+T & SI & scale & 0.0258 & 99.10 & 0.1975 & 84.90 & 0.1050 & 90.89 & 0.4443 & 60.67 \\
\textit{Affine} & & & & S+T & SSI & scale+shift & 0.0234 & 99.16 & 0.1849 & 82.71 & 0.0983 & 90.30 & 0.2959 & 55.59 \\
\textit{Absol} & \checkmark & \checkmark & & S+T & Abs & - & 0.0817 & 95.21 & 0.3319 & 60.50 & 0.2880 & 60.82 & 0.3409 & 37.32 \\
\hline
\textit{CirPad} & \checkmark & & \checkmark & S+T & SI & scale & 0.0258 & 99.10 & 0.0838 & 89.89 & 0.1007 & 90.94 & 0.1934 & 73.35 \\
\hline
\textit{S} & \checkmark & & & S & SI & scale & 0.0238 & 99.16 & 0.4491 & 35.59 & 0.1079 & 88.34 & 0.3377 & 40.51 \\
\textit{SD} & \checkmark & & & S+D & SI & scale & 0.0255 & 99.11 & 0.0719 & 89.31 & 0.1133 & 87.95 & 0.2615 & 58.77 \\
\textit{SDT} & \checkmark & & & S+D+T & SI & scale & 0.0257 & 99.10 & 0.0679 & 92.87 & 0.1007 & 90.98 & 0.1964 & 73.83 \\
\bottomrule
\end{tabular}
}
\caption{Ablation study. Loss: SI---scale-invariant (Eq.~\ref{eq:si_loss}), SSI---scale-and-shift-invariant (MiDaS affine-invariant loss~\cite{ranftl2020towards}), Abs---L1. Datasets: S/D/T---Structured3D/Deep360/TartanAir~v2.}
\label{tab:ablation_study}
\end{table*}

\subsection{Ablation Study}
\label{sec:as}
We conduct ablation studies on the validation/test splits of Structured3D, Deep360, Matterport3D, and Metropolis (Tab.~\ref{tab:ablation_study}). The first block reports \textit{SI}, which is trained with the scale-invariant loss on Structured3D+TartanAir v2 with shift learning but no circular padding; the models in the subsequent blocks are each compared against this reference.

\begin{figure}[t]
\centering
\input{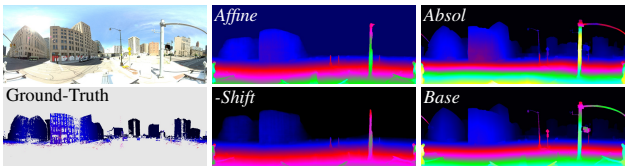}
\caption{Effectiveness of shift learning and loss configuration.}
\label{fig:ablation}
\end{figure}

\textbf{Shift learning and loss configuration.}
We ablate shift learning from \textit{SI} to obtain \textit{NoShift}, and further probe the supervision choice with \textit{Affine} (the scale-and-shift-invariant loss~\cite{ranftl2020towards}, which outputs affine-invariant disparity) and \textit{Absol} (the absolute L1 loss on metric disparity, using two additional MLP heads on the class token to estimate shift and scale). Removing shift learning (\textit{NoShift}) barely changes the indoor Structured3D/Matterport3D scores, but it more than doubles the outdoor error on Deep360/Metropolis, which shows that shift learning is critical for generalization to diverse outdoor scenes. We hypothesize that this is because outdoor scenes (\eg, TartanAir V2) are substantially more varied than indoor ones, so without a learnable shift the model struggles to form scale-invariant representations that generalize across varied environments.

\textit{Affine} relaxes the shift constraint and clearly outperforms \textit{NoShift}, but compared with \textit{SI} it performs notably worse outdoors (about $2{\times}$ the error on Deep360 and $+50\%$ on Metropolis) while only slightly better indoors (11\% lower \textit{AbsRel} on Structured3D and 2\% on Matterport3D)---which indicates that our shift learning is more effective than simply relaxing the shift constraint. \textit{Absol} underperforms \textit{SI} by roughly $2{\times}$ in \textit{AbsRel} on all sets, which suggests that enforcing absolute-depth supervision hurts both in-domain fitting and zero-shot generalization. Fig.~\ref{fig:ablation} shows that only \textit{SI} preserves both near-field detail and far-range magnitude; \textit{Affine}, \textit{Absol}, and \textit{NoShift} all misestimate the magnitudes of distant buildings, which indicates a common failure in capturing accurate depth at large distances.

\textbf{Circular padding.}
\textit{CirPad} yields only minor full-image gains where the borders lack valid depth, but a 5-pixel ERP boundary evaluation (Tab.~\ref{tab:cirp}) reduces the error by over 7\% on Structured3D/Deep360/Metropolis and 2\% on Matterport3D. Fig.~\ref{fig:cirp1} confirms fewer lateral seams and polar artifacts.

\begin{figure}[t]
\centering
\input{figs/cirp}
\vspace{-10pt}
\caption{Comparison of w/o and w/ circular padding.}
\label{fig:cirp1}
\end{figure}

\begin{table}[t]
\centering
\resizebox{0.98\columnwidth}{!}{
\begin{tabular}{l|cccc}
\toprule
Models & Structured3D & Deep360 & Matterport3D & Metropolis \\
\midrule
w/o circular padding & 0.0226 & 0.0958 & 0.1153 & 0.2661 \\
w/ circular padding  & 0.0210 & 0.0871 & 0.1132 & 0.2273 \\
\bottomrule
\end{tabular}
}
\caption{\textit{AbsRel} on a 5-pixel ERP boundary margin.}
\label{tab:cirp}
\end{table}

\textbf{Training datasets.}
The indoor-only \textit{S} improves on the Structured3D validation set but fails to generalize: Matterport3D does not improve, and the outdoor Deep360/Metropolis deteriorate substantially, which shows that indoor data alone is insufficient. Adding Deep360 (\textit{SD}) improves Metropolis by 22\% over \textit{S}, though it remains 33\% worse than \textit{SI}; \textit{SD} also reduces the \textit{AbsRel} by over 10\% compared with PanDA (S) in Tab.~\ref{tab:main} under the same supervised training sets. Using all three sets (\textit{SDT}) achieves strong in-domain fitting but only a marginal improvement over \textit{SI}; given the glass-rendering errors of Deep360, we exclude it from the final training mix to avoid potential negative guidance.

\section{Conclusion}
\label{sec:conclusion}


We present DA360, adapting perspective foundation models for zero-shot panoramic depth estimation. Guided by a comparative analysis identifying DAV2 as the optimal base, we adapt it in disparity space with two lightweight modifications---a learnable shift under scale-invariant supervision and circular padding---yielding seamless 3D point clouds. We further introduce Metropolis, a curated outdoor benchmark of 3{,}000 urban panoramas, on which DA360 achieves SOTA performance. A key limitation is that DA360 predicts scale-invariant rather than metric depth; since naive absolute supervision hurts accuracy (Tab.~\ref{tab:ablation_study}, \textit{Absol}), recovering metric scale without sacrificing generalization remains open.

\fontsize{9.5pt}{10.5pt}
\selectfont
\bibliography{bibliography}

\appendix
\renewcommand{\thesection}{S\arabic{section}}
\renewcommand{\thetable}{S\arabic{table}}
\renewcommand{\thefigure}{S\arabic{figure}}
\setcounter{section}{0}
\setcounter{table}{0}
\setcounter{figure}{0}
\clearpage
\twocolumn[{%
\renewcommand\twocolumn[1][]{#1}%
\begin{center}
{\Large
\textbf{Depth Anything in 360$^\circ$: Towards Scale Invariance in the Wild}\\
\vspace{0.5em}Supplementary Material \\
\vspace{1.0em}
}
\end{center}
\begin{center}
    \captionsetup{type=figure}
    \includegraphics[width=0.95\linewidth]{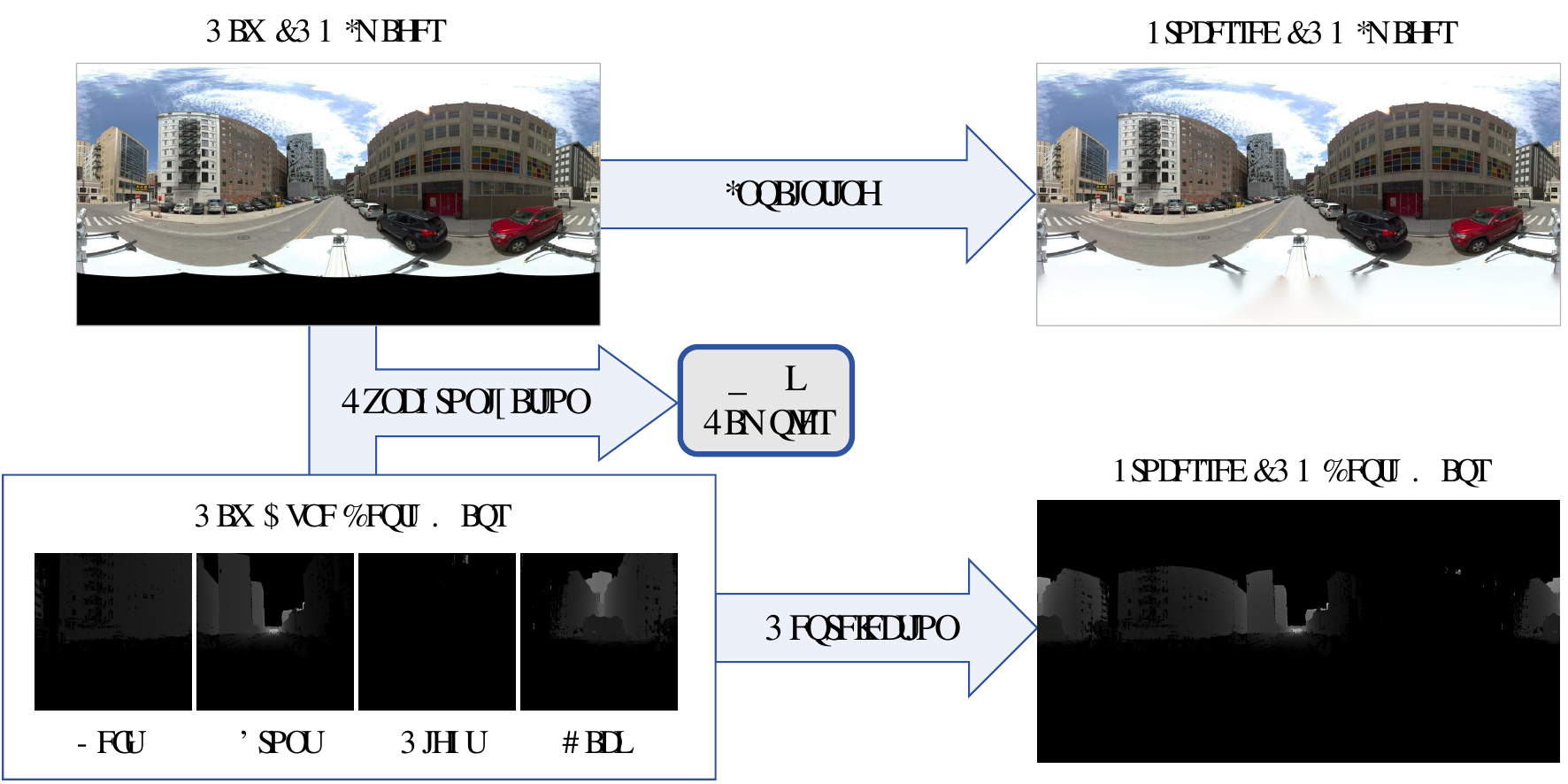}
    \captionof{figure}{Metropolis raw data preprocessing, including sample synchronization, image inpainting and depth map reprojection.}
    \label{fig:metropolits_process}
    \vspace{8pt}
\end{center}%
}]

\begin{figure*}[t]
    \centering{
    \input{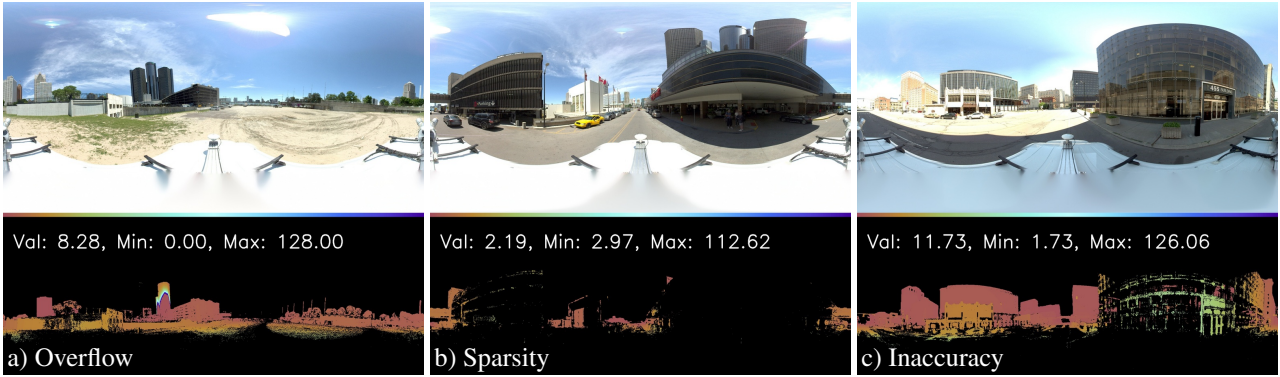}}
    \vspace{-10pt}
   \caption{Typical problematic samples in Metropolis: a) depth value overflow due to encoding limitations; b) overly sparse valid depth; c) apparent depth errors. Val: ratio of valid depth points; Min: minimum depth value; Max: maximum depth value.}
    \label{fig:metropolis2}
    \vspace{13pt}
\end{figure*}

\begin{figure*}[t]
    \centering{
    \input{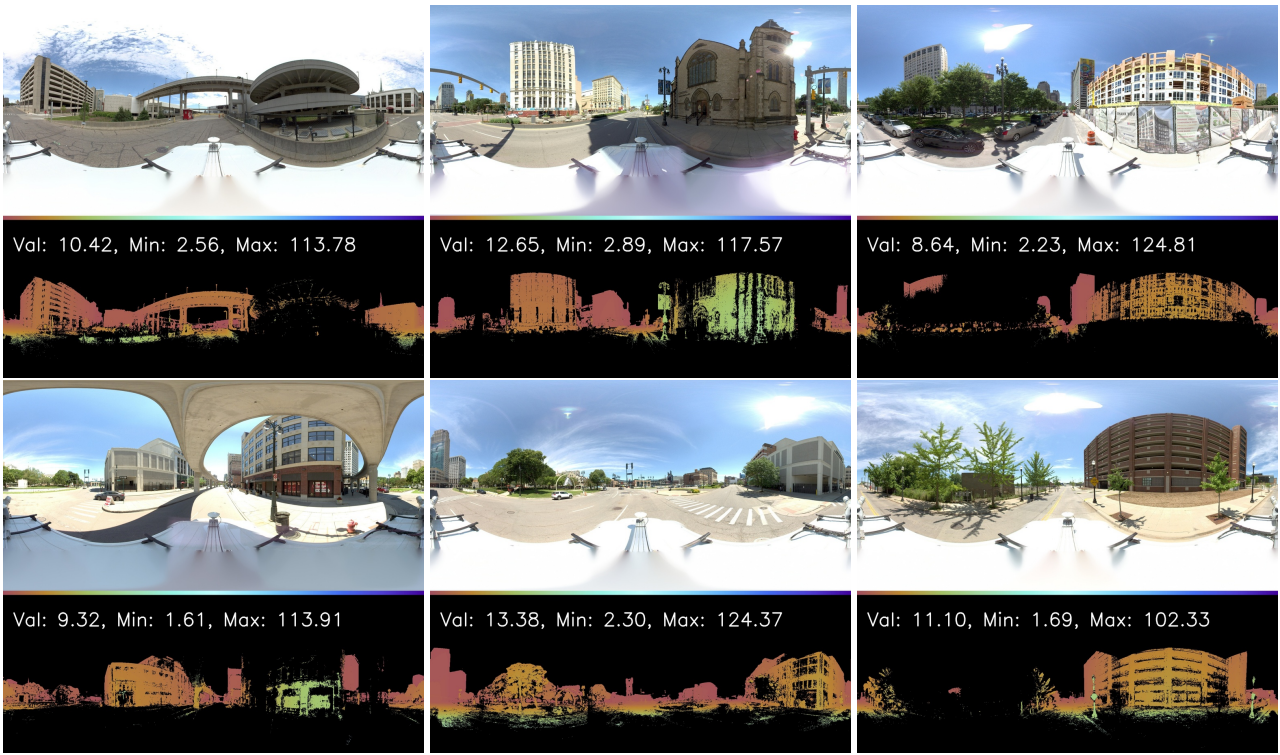}}
    \vspace{-10pt}
    \caption{Six examples of the curated Metropolis test set.}
    \label{fig:metropolis3}
    \vspace{13pt}
\end{figure*}

\section{Metropolis Test Set Construction}
\label{sec:metro_supp}

This section describes our pipeline for curating the raw data from Mapillary Metropolis~\cite{mapillary_metropolis} into a test set for panoramic depth estimation, which consists of two main stages: raw data preprocessing and high-quality sample selection.
The original dataset includes ERP-format panoramas and cubemap Z-buffer depth maps (front, back, left, right), generated by projecting 3D point clouds obtained from aerial/street-level LiDAR, Structure-from-Motion, and Multi-view Stereo reconstructions.
As illustrated in Fig.~\ref{fig:metropolits_process}, the preprocessing stage involves three key steps. Since the five types of data are not perfectly aligned, we first synchronize them, resulting in approximately 10k candidate samples. We then inpaint the empty bottom regions of the ERP panoramas using the bottom-completion method from UniFuse~\cite{jiang2021unifuse} to obtain visually coherent panoramic images. Finally, we convert the cubemap Z-buffer depth maps into ERP-format depth maps, representing the Euclidean distance to the camera center.

After obtaining about 10k panorama-depth pairs through the preprocessing pipeline, visual inspection identified three prevalent issues: (1) depth overflow in certain regions due to limited encoding range; (2) excessively sparse valid depth points; (3) significant depth errors caused by large glass surfaces. Representative examples are illustrated in Fig.~\ref{fig:metropolis2}.

The depth maps in Metropolis are stored as 16-bit integers, with metric depth obtained through the transformation:
\[
d_{\text{metric}} = \frac{d_{\text{int16}}}{256}.
\]
This representation limits the maximum recordable depth to:
\[
d_{\text{max}} = \frac{2^{16} - 1}{256} \approx 255.996 \ \text{meters}.
\]
Values exceeding this threshold wrap around to small numbers during integer conversion, as shown in Fig.~\ref{fig:metropolis2}(a) where building regions beyond 128 meters incorrectly reset to near-zero depth.

To automatically remove samples with depth overflow or sparsity issues, we implemented a filtering script that eliminates samples satisfying either criterion:
\[
\max(d) > 255 \ \text{m} \lor \ \text{ValidRatio}(d) < 8\%.
\]
This process yielded 3.7k candidate samples.

Finally, we conducted manual verification to exclude samples with conspicuous depth errors, resulting in a curated set of 3{,}000 high-quality samples. Six typical samples are shown in Fig.~\ref{fig:metropolis3}, demonstrating diverse urban environments within the dataset.

\section{Additional Experiments and Analyses}
\label{sec:addexp}

\subsection{Disparity Distribution of Training Sets}

This subsection compares the disparity distributions of the two training sets used in our work: TartanAir V2~\cite{patel2025tartanground} (outdoor) and Structured3D~\cite{zheng2020structured3d} (indoor), aiming to highlight their underlying scene differences. For each dataset, we first convert depth to disparity by taking its reciprocal, then normalize each disparity map so that its maximum value equals one. For each normalized disparity map, we compute two statistical measures: the median, which indicates the overall location of the disparity values, and the median absolute deviation (MAD), which reflects their dispersion.

\begin{figure*}[t]
    \centering{
    \includegraphics[width=0.9\linewidth]{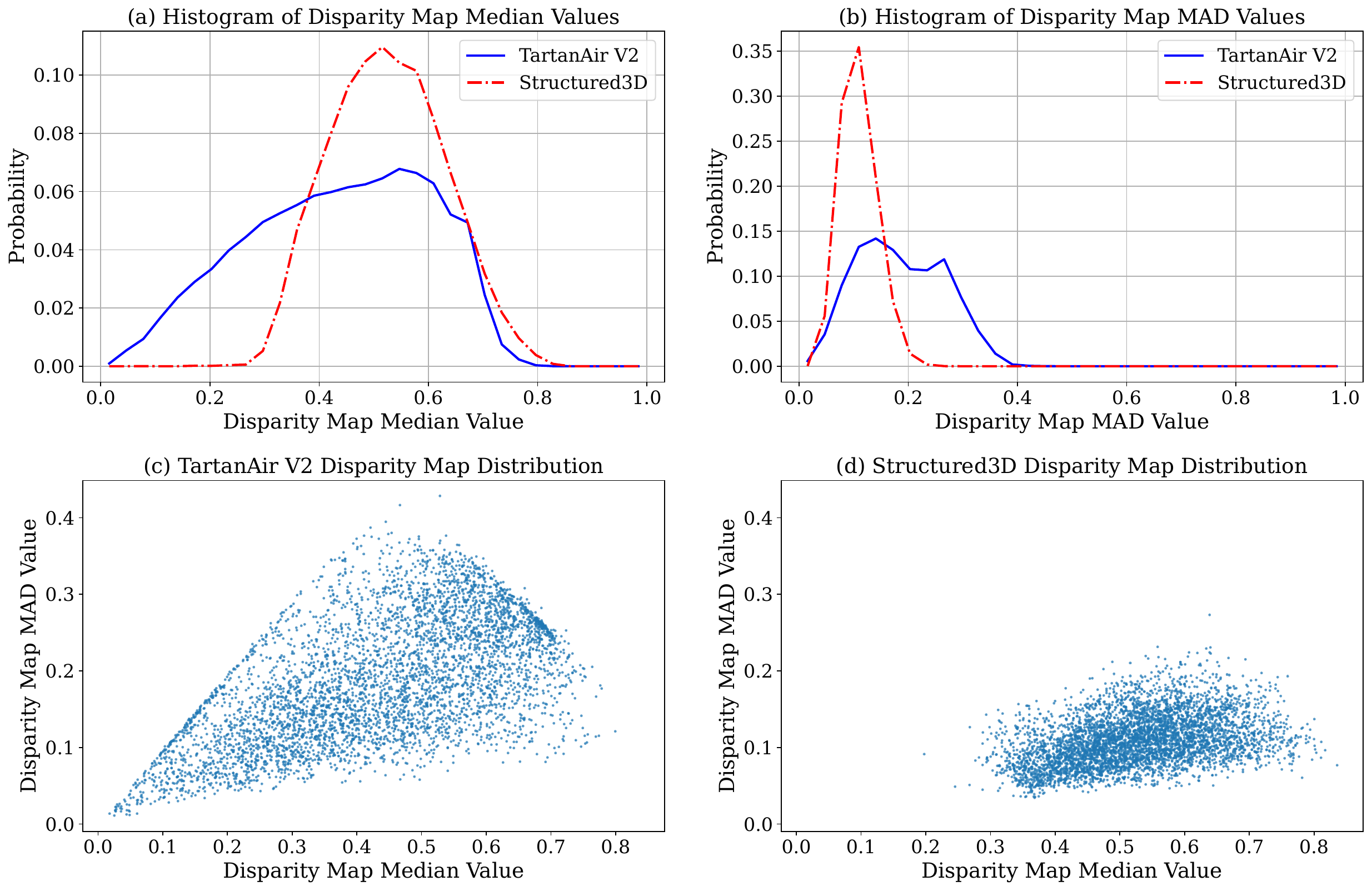}}
    \caption{Comparison of disparity distributions between TartanAir V2~\cite{patel2025tartanground} and Structured3D~\cite{zheng2020structured3d}: (a) histogram of disparity map median values; (b) histogram of disparity map median absolute deviation (MAD) values; (c) and (d) distributions of median and MAD computed from 5,000 randomly selected disparity maps from TartanAir V2 and Structured3D, respectively.}
    \label{fig:disparity}
  \end{figure*}

The comparative results are visualized in Fig.~\ref{fig:disparity}. As shown in Fig.~\ref{fig:disparity} (a), the median disparities of Structured3D are concentrated in the range of 0.3–0.8, with most values lying between 0.4 and 0.7. In contrast, the median disparities of TartanAir V2 span a broader range from 0.0 to 0.8, with a noticeable portion falling in the low region (0.0–0.3). This can be attributed to the presence of distant areas in outdoor scenes, which naturally yield smaller disparity values. Fig.~\ref{fig:disparity} (b) compares the MAD distributions: TartanAir V2 exhibits a significantly wider spread, approximately twice that of Structured3D. Fig.~\ref{fig:disparity} (c) and (d) present scatter plots of median and MAD values for 5,000 randomly sampled disparity maps from each dataset, further confirming that TartanAir V2 occupies a much broader region in the disparity space compared to Structured3D.

These observations indicate that outdoor datasets exhibit substantially greater diversity in disparity statistics than indoor ones, which explains the performance degradation observed when the shift learning module is removed or when only affine-invariant supervision is used, particularly in outdoor generalization.

\begin{figure*}[t]
\centering
\input{supfigs/pretrained}
\caption{Visual comparisons of different model parameter initialization strategies: (\textit{Random}), \textit{DINOv2}, and \textit{DAV2} (corresponding to the model \textit{SI} in the main paper).}
\label{fig:pretrained}
\end{figure*}

\begin{table*}[t]
\centering
\resizebox{2.11\columnwidth}{!}{
  \begin{tabular}{l|c|ccc|ccc|ccc|ccc}
    \toprule
    Model & \multirow{2}{*}{Pretrained Models} &\multicolumn{3}{c|}{Structured3D} &\multicolumn{3}{c|}{Deep360} &\multicolumn{3}{c|}{Matterport3D}  &\multicolumn{3}{c}{Metropolis}\\
    Names &   & \textit{AbsRel}$\downarrow$ & \textit{RMSE}$\downarrow$ & $\delta_1$$\uparrow$ & \textit{AbsRel}$\downarrow$ & \textit{RMSE}$\downarrow$ & $\delta_1$$\uparrow$ & \textit{AbsRel}$\downarrow$ & \textit{RMSE}$\downarrow$ & $\delta_1$$\uparrow$ & \textit{AbsRel}$\downarrow$ & \textit{RMSE}$\downarrow$ & $\delta_1$$\uparrow$ \\
    \midrule
    \textit{DAV2} & Depth Anything V2~\cite{yang2024depth} & 0.0263 & 0.0988 & 99.07 & 0.0879 & 6.2732 & 89.34 & 0.1004 & 0.5047 & 91.02 & 0.1966 & 14.587 & 72.78 \\
    \textit{DINOv2} & DINOv2~\cite{oquab2023dinov2} & 0.0348 & 0.1218 & 98.36 & 0.0967 & 6.9283 & 87.60 & 0.1158 & 0.5611 & 88.09 & 0.2421 & 17.059 & 62.97 \\
    \textit{Random} & None/Random & 0.1444 & 0.3278 & 82.01 & 0.5039 & 29.6632 & 37.06 & 0.3000 & 1.0870 & 50.83 & 0.5266 & 17.675 & 32.57 \\
    \bottomrule
  \end{tabular}
    }
  \caption{Ablation study on different model parameter initialization strategies.}
  \label{tab:pretrained}
\end{table*}

\subsection{Random vs. DINOv2 vs. DAV2 Initialization}

This section further validates the effectiveness of using Depth Anything V2 (DAV2)~\cite{yang2024depth} for parameter initialization. Building upon the \textit{SI} (here denoted as \textit{DAV2}) model from ablation studies in the main paper, we introduce two additional variants: one initialized with DINOv2~\cite{oquab2023dinov2} and another trained from scratch with random initialization. As summarized in Tab.~\ref{tab:pretrained}, all three models share identical configurations except for their initialization strategies.

The results clearly demonstrate that omitting DAV2 pretraining leads to noticeable performance degradation across all evaluation metrics. While using the vision foundation model DINOv2 yields a relatively smaller decline---with relative depth error increasing by 10\%--30\% due to its strong semantic pretraining---the degradation remains substantial, particularly in fine-grained geometric details. Training from scratch produces unacceptable outcomes, where the relative error increases by 270\%--570\%, indicating that the model fails to learn reliable depth representations from limited panoramic training data alone.

Fig.~\ref{fig:pretrained} provides a visual comparison. Random initialization results in coarse and unreliable depth estimates. DINOv2 initialization captures rough scene structures but lacks geometric precision. In contrast, DAV2 initialization produces accurate and detailed depth maps. These results demonstrate the distinct advantage of leveraging DAV2 for pretraining, particularly when annotated panoramic depth data is scarce.

\begin{table*}
\centering
\vspace{5pt}
\resizebox{2.0\columnwidth}{!}{
  \begin{tabular}{l|c|c|c|ccc|ccc|ccc}
    \toprule
    \multirow{2}{*}{Models} & Pre-trained & Finetuning & Evaluation &\multicolumn{3}{c|}{Matterport3D} &\multicolumn{3}{c|}{Stanford2D3D} &\multicolumn{3}{c}{Metropolis}\\
    & Models & Methods & Alignment & \textit{AbsRel}$\downarrow$ & \textit{RMSE}$\downarrow$ & $\delta_1\uparrow$ & \textit{AbsRel}$\downarrow$ & \textit{RMSE}$\downarrow$ & $\delta_1\uparrow$ & \textit{AbsRel}$\downarrow$ & \textit{RMSE}$\downarrow$ & $\delta_1\uparrow$ \\
    \midrule
    PanDA(S)* & DAV2-ind & PanDA & scale+shift & 0.1346 & 0.5090 & 83.59 & 0.0957 & 0.3261 & 90.80 & 0.2898 & 15.037 & 57.52 \\
    PanDA(S)$^\dag$ & DAV2-out & PanDA & scale+shift & 0.1414 & 0.5328 & 83.26 & 0.0926 & 0.3400 & 92.87 & 0.2994 & 15.362 & 56.90 \\
    PanDA(S)$^\circ$ & DAV2 & PanDA & scale+shift & 0.1319 & 0.5028 & 85.20 & 0.0930 & 0.3293 & 91.75 & 0.2517 & 14.197 & 64.15 \\
    \hline
    DA360(S)* & DAV2-ind & DA360 & scale & 0.1076 & 0.5780 & 89.48 & 0.0754 & 0.3569 & 93.95 & 0.1990 & 14.673 & 72.77 \\
    DA360(S)$^\dag$ & DAV2-out & DA360 & scale &  0.1059 & 0.5358 & 90.33 & 0.0739  & 0.3210 & 94.73 & 0.2465 & 15.065 & 67.02 \\
    DA360(S) & DAV2 & DA360 & scale & 0.0963 & 0.4971 & 91.60 & 0.0687 & 0.3009 & 94.84 & 0.1922 & 14.286 & 74.22 \\
    \bottomrule
  \end{tabular}
    }
  \caption{Different pre-trained models and finetuning methods.}
  \label{tab:pretrain_finetune}
\end{table*}

\subsection{Pretraining and Finetuning Combinations}
\label{sec:pretrain_and_finetune}

In this subsection, we directly compare different DAV2 variants as pretrained backbones, fine-tuned with two distinct supervision strategies: PanDA's depth-space optimization and our DA360's disparity-space adaptation. All experiments use the ViT-Small backbone and follow the same training protocol as the main experiments, i.e., 300k iterations on Structured3D and TartanAir v2 with identical learning rate schedules and optimizers.

For PanDA-style fine-tuning on DAV2 models, whose output is disparity, we invert the sign of the final layer's weights and bias (multiplying by \(-1\)) to produce a depth-ordered output after sigmoid activation, facilitating optimization in depth space. Similarly, when applying DA360 to DAV2-ind and DAV2-out, we also negate the final layer parameters to preserve disparity-compatible output ordering.

Results are presented in Table~\ref{tab:pretrain_finetune}, where PanDA(S)* and DA360(S) are identical to their counterparts in the main table. Across all fine-tuning methods, the vanilla DAV2 consistently outperforms both metric-depth variants, confirming that it serves as a more effective starting point. Moreover, our DA360 fine-tuning approach consistently and substantially outperforms PanDA across all pretrained models, demonstrating the clear advantage of our adaptation strategy.

\subsection{Resolution Scalability Analysis}
\label{sec:resolution}

Panoramic depth estimation methods are typically evaluated at resolutions around $512\times1024$, while modern cameras can capture images at much higher resolutions (e.g., $1024\times2048$). A practical solution should handle such inputs without significant performance degradation. We analyze the resolution scalability of our trained DA360 models to address this gap. We evaluate three DA360 variants (ViT-S, ViT-B, ViT-L) on Matterport3D (indoor) and Metropolis (outdoor). All models are trained at $1036\times518$. We compare three protocols: \begin{itemize}
\item \textbf{1$\times$}: Standard evaluation at training resolution ($518\times1036$).
\item \textbf{2$\times$}: Direct inference on images upsampled to $1036\times2072$.
\item \textbf{1$\times \to$ 2$\times$}: Inference at $1\times$ followed by bilinear upsampling of the predicted depth map to $2\times$.
\end{itemize}

\begin{table}
\centering
\resizebox{0.99\columnwidth}{!}{
  \begin{tabular}{l|ccc|ccc}
    \toprule
    \multirow{2}{*}{Models} &\multicolumn{3}{c|}{Matterport3D}  &\multicolumn{3}{c}{Metropolis}\\
     & 1$\times$  & 2$\times$ & 1$\times\to$2$\times$ & 1$\times$  & 2$\times$ & 1$\times\to$2$\times$  \\
    \midrule
    \bf{DA360 (S)} & 0.0963 & 0.1259 & 0.0948 & 0.1922 & 0.2332 & 0.1919 \\
    \bf{DA360 (B)} & 0.0877 & 0.1209 & 0.0861 & 0.1798 & 0.2269 & 0.1783 \\
    \bf{DA360 (L)} & 0.0793 & 0.1075 & 0.0783 & 0.1630 & 0.2054 & 0.1622 \\
    \bottomrule
  \end{tabular}
    }
    \vspace{-6pt}
  \caption{$AbsRel$ at different resolution settings.}
  \label{tab:resolution_scalability}
\end{table}

Tab.~\ref{tab:resolution_scalability} reports absolute relative error (\textit{AbsRel}). Direct inference at $2\times$ causes a 20–30\% performance drop, confirming the cross-resolution generalization challenge. Notably, the \textbf{1$\times \to$ 2$\times$} pipeline consistently matches or exceeds native $1\times$ performance. For instance, DA360 (L) achieves 0.1630 ($1\times$), 0.2054 ($2\times$), and 0.1622 ($1\times \to 2\times$) on Metropolis.

We attribute this to the low-frequency nature of depth maps: upsampling preserves structural information without introducing artifacts. In contrast, feeding high-resolution RGB images directly may introduce high-frequency details unseen during training, confusing the predictor. This indicates that our model learns robust geometric representations at low resolution that transfer effectively to higher resolutions through simple upsampling. For practical deployment requiring high-resolution depth, running the model at its native training resolution followed by output upsampling offers a simple, cost-effective solution without retraining. We include this analysis to demonstrate our method's robustness and provide guidance for real-world applications.

\begin{table*}
\centering
\vspace{5pt}
\resizebox{2.0\columnwidth}{!}{
  \begin{tabular}{l|c|c|c|ccc|ccc|ccc}
    \toprule
    \multirow{2}{*}{Models} & Inference & Run & Evaluation &\multicolumn{3}{c|}{Matterport3D} &\multicolumn{3}{c|}{Stanford2D3D} &\multicolumn{3}{c}{Metropolis}\\
    & Style & Time & Alignment & \textit{AbsRel}$\downarrow$ & \textit{RMSE}$\downarrow$ & $\delta_1\uparrow$ & \textit{AbsRel}$\downarrow$ & \textit{RMSE}$\downarrow$ & $\delta_1\uparrow$ & \textit{AbsRel}$\downarrow$ & \textit{RMSE}$\downarrow$ & $\delta_1\uparrow$ \\
    \midrule
    MoGe-2(L) & end2end & 0.101s & scale+shift & 0.2920  & 0.7827 & 54.65 & 0.3731 & 0.9519 & 45.65 & 0.6764 & 23.152 & 30.84 \\
    MoGe-2(L) & crop\&fuse & 196s & scale+shift & 0.0835 & 0.3519 & 93.89 & 0.1134 & 0.2984 & 86.15 & 0.2141 & 12.644 & 72.04 \\
    DAV2(L) & crop\&fuse & 152s & scale+shift & 0.1195 & 0.5704 & 86.34 & 0.1491 & 0.5633 & 78.90 & 0.2230 & 16.201 & 62.66 \\
    \hline
    DA360(L) & end2end & 0.026s & scale & 0.0793 & 0.4407 & 94.07 & 0.0710 & 0.2930 & 93.53 & 0.1630 & 13.015 & 80.47 \\
    \bottomrule
  \end{tabular}
    }
  \vspace{-8pt}
  \caption{Comparison with crop\&fuse baselines.}
  \label{tab:crop_fuse}
  \vspace{0pt}
\end{table*}

\subsection{Additional Quantitative Comparison}

\textbf{Comparison with crop\&fuse baselines.}
We further compare DA360 against \emph{crop\&fuse} pipelines, which estimate depth on multiple perspective (tangent) crops of the panorama and merge the predictions through an optimization-based fusion stage. Results are reported in Tab.~\ref{tab:crop_fuse}. Feeding a perspective model such as MoGe-2 directly with an ERP panorama (\textit{end2end}) collapses—\textit{AbsRel} degrades to 0.2920/0.3731/0.6764 on Matterport3D/Stanford2D3D/Metropolis—confirming that perspective models cannot cope with panoramic distortion out of the box. The crop\&fuse strategy largely restores accuracy for both MoGe-2 and DAV2, but at a prohibitive cost (196s and 152s per image). In contrast, our DA360 (L) requires only a single end-to-end forward pass (0.026s) yet still surpasses both crop\&fuse baselines on \emph{every} metric across all three datasets; for instance, on Metropolis it lowers \textit{AbsRel} from 0.2141 (MoGe-2, crop\&fuse) to 0.1630 and raises $\delta_1$ from 72.04 to 80.47. This demonstrates that DA360 attains crop\&fuse-level (or better) accuracy at roughly four orders of magnitude lower latency, without any test-time optimization.

\textbf{Running time.}
The running time of end-to-end models differs drastically from that of post-fusion methods. Inference time was measured on a system with 12 CPU cores and an NVIDIA RTX 4090 GPU, averaged over 100 samples. End-to-end models are fast, ranging from roughly 0.02s to 0.06s. DAV2 is the fastest at 0.024s. Our DA360, which adds only a minimal MLP for shift estimation and employs circular padding, requires 0.026s, while PanDA, incorporating LoRA~\cite{hu2022lora} layers into the ViT backbone, takes 0.028s. The other two end-to-end models, UniK3D and DA$^{2}$, feature custom modules and require 0.053s and 0.031s, respectively. In stark contrast, the post-fusion methods MoGe~\cite{wang2025moge} and MoGe-2~\cite{wang2025moge2} incur substantial costs, demanding 124s and 196s. Applying the same crop\&fuse pipeline to DAV2~\cite{yang2024depth} likewise incurs a heavy cost of 152s per image. Within this budget, depth estimation on the 12 tangent images (resolution $512\times512$) using networks of comparable size accounts for only a minor fraction of the total time; the optimization-based fusion stage dominates, highlighting the significant computational burden inherent to fusion strategies.

\begin{figure*}
    \centering{
    \includegraphics[width=1.0\linewidth]{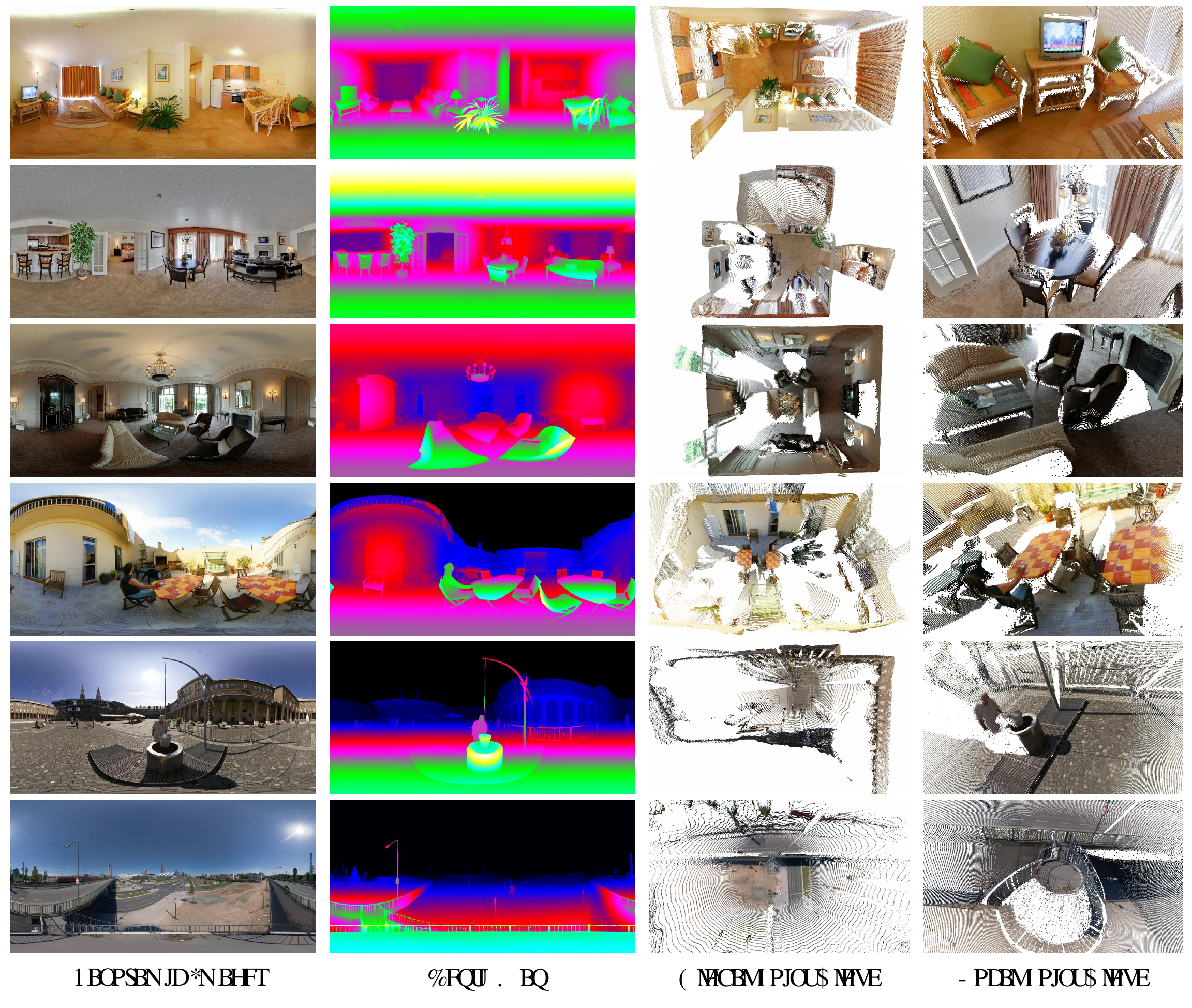}}
    \vspace{-10pt}
    \caption{Additional visualizations on SUN360: six examples with depth maps and point clouds from our DA360.}
    \label{fig:addition}
    \vspace{-10pt}
\end{figure*}

\newpage

\subsection{Additional Qualitative Results}
\label{sec:aqr}

This section presents extended qualitative evaluations on the diverse SUN360 panoramic dataset~\cite{xiao2012recognizing}. Fig.~\ref{fig:addition} presents six additional qualitative examples from the SUN360 dataset, showcasing DA360's predicted depth maps, overall 3D point cloud renderings, and close-up views of local regions. The visualization further demonstrates the accuracy of DA360. There is an attached video file (ODV360.mp4) that provides additional visual comparisons of 20 diverse video clips from the ODV360 dataset~\cite{cao2023ntire}.

We include more comparative examples of estimated depth maps against several relevant methods: DAV2~\cite{yang2024depth} and PanDA~\cite{cao2025panda} as direct competitors; UniK3D~\cite{piccinelli2025unik3d} for arbitrary camera types; DA$^{2}$~\cite{li2025depth} as a concurrent scale-invariant panoramic depth model and MoGe~\cite{wang2025moge}/MoGe-2~\cite{wang2025moge2} which employ tangent plane projection with optimization-based fusion. All methods process the panoramic images at $518\times1036$ resolution and use the ViT-L backbone in their neural models.

Ten examples are presented in Fig.~\ref{fig:in1}-\ref{fig:out5}, respectively. Our DA360 model demonstrates superior depth estimation accuracy with more precise global scales and finer local details. Furthermore, we observe that fusion-based methods like MoGe may fail to seamlessly integrate depth predictions from multiple tangent planes, as shown in the white rectangular region of example \uppercase\expandafter{\romannumeral 1}. 
Among all baseline models, the MoGe series, particularly MoGe-2, recovers finer details. This benefit stems from its higher effective input resolution, i.e., twelve $512\times512$ tangent images. However, this comes at the cost of requiring time-consuming post-processing. Remarkably, even at a single $518\times1036$ input, our DA360 produces excellent detail, matching MoGe-2's quality in most cases and even surpassing it in some. This clearly demonstrates the superiority of DA360.

\newpage

\begin{figure*}[t]
\centering
\input{supfigs/comparisons/indoor1}
\vspace{-5pt}
\caption{Visual comparison on example \uppercase\expandafter{\romannumeral 1} of SUN360~\cite{xiao2012recognizing} with SOTA depth estimation models. }
\label{fig:in1}
\end{figure*}

\newpage

\begin{figure*}[t]
\centering
\input{supfigs/comparisons/indoor2}
\vspace{-5pt}
\caption{Visual comparison on example \uppercase\expandafter{\romannumeral 2} of SUN360~\cite{xiao2012recognizing} with SOTA depth estimation models. }
\label{fig:in2}
\end{figure*}

\newpage

\begin{figure*}[t]
\centering
\input{supfigs/comparisons/indoor3}
\vspace{-5pt}
\caption{Visual comparison on example \uppercase\expandafter{\romannumeral 3} of SUN360~\cite{xiao2012recognizing} with SOTA depth estimation models. }
\label{fig:in3}
\end{figure*}

\newpage

\begin{figure*}[t]
\centering
\input{supfigs/comparisons/indoor4}
\vspace{-5pt}
\caption{Visual comparison on example \uppercase\expandafter{\romannumeral 4} of SUN360~\cite{xiao2012recognizing} with SOTA depth estimation models. }
\label{fig:in4}
\end{figure*}

\newpage

\begin{figure*}[t]
\centering
\input{supfigs/comparisons/indoor5}
\vspace{-5pt}
\caption{Visual comparison on example \uppercase\expandafter{\romannumeral 5} of SUN360~\cite{xiao2012recognizing} with SOTA depth estimation models. }
\label{fig:in5}
\end{figure*}

\newpage

\begin{figure*}[t]
\centering
\input{supfigs/comparisons/outdoor1}
\vspace{-5pt}
\caption{Visual comparison on example \uppercase\expandafter{\romannumeral 6} of SUN360~\cite{xiao2012recognizing} with SOTA depth estimation models. }
\label{fig:out1}
\end{figure*}

\begin{figure*}[t]
\centering
\input{supfigs/comparisons/outdoor2}
\vspace{-5pt}
\caption{Visual comparison on example \uppercase\expandafter{\romannumeral 7} of SUN360~\cite{xiao2012recognizing} with SOTA depth estimation models. }
\label{fig:out2}
\end{figure*}

\begin{figure*}[t]
\centering
\input{supfigs/comparisons/outdoor3}
\vspace{-5pt}
\caption{Visual comparison on example \uppercase\expandafter{\romannumeral 8} of SUN360~\cite{xiao2012recognizing} with SOTA depth estimation models. }
\label{fig:out3}
\end{figure*}

\begin{figure*}[t]
\centering
\input{supfigs/comparisons/outdoor4}
\vspace{-5pt}
\caption{Visual comparison on example \uppercase\expandafter{\romannumeral 9} of SUN360~\cite{xiao2012recognizing} with SOTA depth estimation models. }
\label{fig:out4}
\end{figure*}

\begin{figure*}[t]
\centering
\input{supfigs/comparisons/outdoor5}
\vspace{-5pt}
\caption{Visual comparison on example \uppercase\expandafter{\romannumeral 10} of SUN360~\cite{xiao2012recognizing} with SOTA depth estimation models. }
\label{fig:out5}
\end{figure*}

\end{document}